**World Trade Center responders in their own words:**
**Predicting PTSD symptom trajectories with AI-based language analyses of interviews**
Running head: PTSD Symptom Trajectories from AI-Based Language Analysis

Youngseo Son[1]
Sean A. P. Clouston[3,4]
Roman Kotov[2]
Johannes C. Eichstaedt[6]
Evelyn J. Bromet[2]
Benjamin J. Luft[5]
H Andrew Schwartz[1]

[1]Department of Computer Science, Stony Brook University, Stony Brook, New York, USA
[2]Department of Psychiatry, Stony Brook University, Stony Brook, New York, USA
[3]Program in Public Health, Stony Brook University, Stony Brook, New York, USA
[4]Department of Family, Population and Preventive Medicine, Stony Brook University, Stony Brook, New York, USA
[5]Department of Medicine, Stony Brook University, Stony Brook, New York, USA
[6]Department of Psychology & Institute for Human-Centered A.I., Stanford University, Stanford, California, USA

Corresponding author:
Youngseo Son, Ph.D. Candidate
Departments of Computer Science
Stony Brook University
yson@cs.stonybrook.edu

This work was supported by, in part, NIH R01 AA028032-01.

**Abstract**

**Background:** Oral histories from 9/11 responders to the World Trade Center (WTC) attacks provide rich narratives about distress and resilience. Artificial Intelligence (AI) models promise to detect psychopathology in natural language, but they have been evaluated primarily in non-clinical settings using social media. This study sought to test the ability of AI-based language assessments to predict PTSD symptom trajectories among responders.

**Methods:** Participants were 124 responders whose health was monitored at the Stony Brook WTC Health and Wellness Program who completed oral history interviews about their initial WTC experiences. PTSD symptom severity was measured longitudinally using the PTSD Checklist (PCL) for up to 7 years post-interview. AI-based indicators were computed for depression, anxiety, neuroticism, and extraversion along with dictionary-based measures of linguistic and interpersonal style. Linear regression and multilevel models estimated associations of AI indicators with concurrent and subsequent PTSD symptom severity (significance adjusted by false discovery rate).

**Results:** Cross-sectionally, greater depressive language ($\beta$=0.32; p=0.043) and first-person singular usage ($\beta$=0.31; p=0.044) were associated with increased symptom severity. Longitudinally, anxious language predicted future worsening in PCL scores ($\beta$=0.31; p=0.031), whereas first-person plural usage ($\beta$=-0.37; p=0.007) and longer words usage ($\beta$=-0.36; p=0.007) predicted improvement.

**Conclusions:** This is the first study to demonstrate the value of AI in understanding PTSD in a vulnerable population. Future studies should extend this application to other trauma exposures and to other demographic groups, especially under-represented minorities.

*Key words:*

9/11, World Trade Center, disaster responders, oral history interviews, language-based assessments, posttraumatic stress disorder, depression, trajectories, risk factors

**Introduction**

The 9/11 attacks on the World Trade Center (WTC) left thousands of casualties and drastically affected the lives of hundreds of thousands of New Yorkers and others nearby (Bergen, 2019). Many affected were those dedicating their lives to the safety of others -- police, firefighters, emergency medical personnel, and other responders to the crisis. There has been a large physical and mental burden of the events that day which has left many struggling with their health as they age (Durkin, 2018; Luft et al., 2012). Many responders suffer from PTSD which has been either worsening, staying the same, or gradually improving over time (Cukor et al., 2011; Neria et al., 2010).

Huge disasters, such as the WTC attacks, have the potential to affect a large number of people at the same time and usually occur within a relatively short period of time. Illuminating the risk and protective factors that reliably predict future reductions or increases in PTSD symptoms can lead to improved understanding, easier in-clinic guidance on patient's well-being, and more rapid care for those involved in catastrophic events. Previous work has made major headway in establishing longitudinal associations of exposure severity, demographic characteristics, and job duties with health trajectories of WTC responders (Cone et al., 2015; Bromet et al., 2016; Pietrzak et al,. 2014). However, additional approaches to risk assessment are needed to more rapidly and fully differentiate those at greatest risk in situations where structured approaches to data collection are not possible. Recently, AI-based techniques have begun to show promise for quickly and accurately assessing mental health from human behavioral data, such as language use patterns. For example, from social media language, researchers have predicted those more prone to post-partum depression (De Choudhury, Kiciman, Dredze, Coppersmith, & Kumar, 2016), those more likely to receive a clinical diagnosis of depression (Eichstaedt et al., 2018) or those appearing at greatest risk of suicide (Zirikly, Resnik, Uzuner, & Hollingshead, 2019; Matero et al., 2019). In all such cases, modern machine learning techniques are used to automatically extract and quantify patterns of language from hundreds to thousands of words per individual, which are then used to automatically produce a mental health or risk score. As compared to traditional questionnaire-based assessments, such approaches seem to suffer from fewer self-report biases (Youyou, Kosinski, & Stillwell, 2015) and generally leverage a larger amount of information per person (Kern et al., 2016). However, using such approaches in a clinical setting not only requires willingness of patients to share private information from social media pages, but also requires that each participant has a substantial amount of data to share in the first place.

In this study, we present the first evaluation of AI-based mental health assessments from language (henceforth language-based assessments) for use in predicting future PTSD symptom trajectories of patients monitored in a clinical setting. Rather than social media, we utilize transcripts of oral history interviews from responders to the 9/11 attacks. We first examine whether existing ("pre-trained") predictive models (most of which were trained on social media) produce assessments associated with PTSD symptoms scores close to the time of interview. We then compare these language assessments to other information available within a mental health clinical cohort (e.g., age, gender, occupation) in order to evaluate the additional benefit of the AI-based assessments. Lastly, we seek to quantify the predictive power of language based indicators, in part to assess their potential suitability for informing personalized therapeutic approaches.

**Methods**

*Participants*

The sample was derived from Stony Brook University's World Trade Center (WTC) Health & Wellness program, funded by the Centers for Disease Control and Prevention, that provides ongoing monitoring of WTC responders. A total of *N*=124 responders underwent an oral history interview and agreed to allow researchers to merge data from the transcript of the oral history with information in their health monitoring records. Hammock et al. (2019) provide an extensive summary of data collection methods. Briefly, oral history participants were primarily recruited *via* word of mouth and by flyers posted in the Stony Brook WTC Wellness Program.

Each interview lasted approximately one hour. It covered the responders' memory of 9/11 attacks and disaster relief efforts, their work activities at the site, experiences and sensations over the days and weeks that followed, and how the WTC disaster ultimately impacted their lives since. Interviews were conducted by clinical staff with diverse healthcare backgrounds after a comprehensive orientation in conducting guided interviews and eliciting details relevant to the key topics to be covered.. Responders were encouraged to discuss what was most important to them. Interviews were completed between 2010 and 2018.

In order to restrict our sample responders who were not new to the WTC Health Program, the analysis sample was restricted to participants who had at least one valid score on the PTSD Checklist (PCL; Blanchard, Jones-Alexander, Buckley, & Forneris, 1996) within two years of their interview, and at least one pre-interview PCL yielding an analysis sample of $N=113$ responders. The few newer health program enrollees this excluded were qualitatively different, having only had just begun care (and potential PTSD treatment) at interview time. Furthermore, to study longitudinal trajectories post-interview, we focused on the subset of individuals with at least three post-interview mental health assessments at least two years after the interview (N=75). The demographic characteristics of the study samples are listed in Table 1. The demographic ratio of gender and police remained similar (<4% difference) after we limited the sample to responders who met criteria for theour language analysis; 92% of the subset group were male and 49% were police; their mean age at interview was 53.

*Ethics:* This study was approved by the Stony Brook University Institutional Review Board. The participants provided written informed consent.

**Table 1.** Data on subjects for health state correlation cross-sectional analysis and trajectory predictions

|  | N | Female | Police | Mean age at the interview (SD) | Median number of words |
|---|---|---|---|---|---|
| All Participants | 124 | 10% | 48% | 55.4 (9.8) | 10,254 |
| Meet Inclusion Criteria | 75 | 8% | 49% | 53.4 (9.5) | 9,944 |

*Language-based assessments*

We automatically derived nine variables assessing the responders language during the interviews: four AI-based assessments of psychological traits (expression of anxiety, depression, neuroticism, and extraversion), three lexicon-based assessments of language style (first-person singular pronouns, plural pronouns, and use of articles), and two meta variables describing counts of words and lengths of words. The process to get these variables consisted of three steps: text transcription, conversion of text to linguistic features, and application of AI based models or *lexica*.

Audio of each interview was transcribed into text using *TranscribeMe,* a HIPAA-approved transcription service. Each time the responders spoke, transcribers labelled the time and the words mentioned. The text of each interview was converted into "features" --- quantitative values describing the content of the interview language -- and then input into: (a) four AI-based assessments of psychological traits, (b) three lexicon-based assessments of language style, and (c) two meta-variable extractions describing counts of words and lengths of words. All analyses, described below, were performed using the Differential Language Analysis ToolKit (DLATK) (Schwartz et al., 2017).

*Conversion into Linguistic Features.*

The models we used required up to three types of linguistic features: (1) relative frequencies of words and phrases, (2) binary indicators of words and phrases, and (3) topic prevalence scores. Words and phrases are sequences of 1 to 3 words in a row. Their relative frequency was recorded by *DLATK* by counting each word or phrases mentioned and dividing by the total number of words or phrases mentioned by the responder. The binary indicator for words and phrases simply indicated whether each word or phrase shows up (1) or not (0). The tokenizer built into the *DLATK* package was used to extract words per interview.

Topics are weighted groups of semantically-related words, often derived through a statistical process called latent dirichlet allocation (LDA) (Blei, Ng, & Jordan, 2003). Once derived, topics can be applied to textual data to scoring, ranging from 0 to 1, indicating how frequently each group of words was mentioned (Kern et al., 2016). We use a standard set of 2,000 topics introduced by Schwartz et al. (2013), which has frequently been applied in the psychological domain including most recently in (Eichstaedt et al., 2020). Once extracted, features were mapped to nine coarse-grained scores as described below and used for analyses herein.

*AI-Based Psychological Traits (4)*

The AI-based assessments input linguistic features such as words, phrases, and topics, and map them to psychological constructs (Kern et al., 2016; Schwartz & Ungar, 2015). We focused on existing pre-trained models for constructs known to be related to our mental health outcomes: (1) neuroticism and (2) extraversion (Park et al., 2015; Schwartz et al., 2013) - the two factors of the five factor model known to relate negatively to depression and anxiety-related mental health conditions (Jorm et al., 2000; Farmer et al., 2002; Jylhä and Isometsä, 2006), as well as (3) degree of depression and (4) anxiousness (Schwartz et al., 2014) - subfacets of emotional stability which correspond to negative high arousal language (anxiousness) and negative low arousal language (depressive). These models were trained on large and diverse populations (approximately sample sizes of N = 65,000 for neuroticism and extraversion and N = 29,000 for degrees of depression and anxiousness). They utilize the linguistic features previously mentioned words and phrases as well as topics as input and output continuous scores for each of the four constructs. They have been validated against standard questionnaire-based measures as well as convergent factors and external criteria under a range of situations (Schwartz et al., 2014; Park et al., 2015; Kern et al., 2016; Matero et al., 2019). However, the predictive validity of these models has yet to be assessed in clinical interview settings.

*Function Word Lexicon Features (3)*

We extracted word frequencies of terms in LIWC 2015 categories (Pennebaker et al., 2015) and calculated categories for an interview with each responder. Due to the relatively low sample size, we focused on the function word categories which were most prevalent and then selected those that had a literature suggested association with mental health:

- First-person singular: depressed, low status, personal, emotional, informal. Previously correlated positively with neuroticism, depression and anxiety (Baddeley & Singer, 2008; Rude, Gortner, & Pennebaker, 2004; Holtzman, 2017) and negatively with life satisfaction (Schwartz et al., 2013).
- First-person plural: high status, socially connected to group. Previously correlated negatively with depression and anxiety (Ramirez-Esparza, Chung, Kacewicz, & Pennebaker, 2008) and positively correlated with life satisfaction (Schwartz et al., 2013) along with the cognition and psychological well-being variables of our interest (Tausczik & Pennebaker, 2010):
- Articles: use of concrete nouns, interest in objects and things (Tausczik & Pennebaker, 2010).

*Language meta features (2)*

- Average word length feature: high cognitive load (Khawaja, Chen, & Marcus, 2010; Lewis and Frank, 2016), education, and social class (Hartley, Pennebaker, & Fox, 2003; Tausczik and Pennebaker, 2010)

- Word counts: We also recorded total word counts, the number by which all lexica above were normalized. Given the interviews were all an hour long, this is a proxy for rate of speech from each participant.

*Mental health outcomes*

The PTSD Symptom Checklist for DSM-IV PTSD (PCL) was used to assess PTSD severity in the past month (Cone et al., 2015; Bromet et al., 2016; Pietrzak et al., 2014). We chose the PCL closest to the interview date (all within two years) for concurrent analyses (average initial PCL score = 33.7; SD = 16.2). Following previous work which suggests that a fixed cutoff might not be optimally established for all cases (Andrykowski, Cordova, Studts, & Miller, 1998; Bovin et al., 2016), we focused on continuous values. Post-interview PCL scores were used to create trajectories as described under *trajectory prediction* below.

*Statistical analysis*

We used linear regression coefficient of the target explanatory variable (PCL score) as its correlation strength and multivariable adjustment for possible confounders (age, gender, occupation and years after 9/11) to acquire the unique effects of language-based assessments. Since we explored many language assessments at once, we considered coefficients significant if their Benjamini-Hochburg adjusted p-values were less than 0.05.

*Concurrent Evaluation*

We processed the interviews of responders who had PTSD assessments three or more interviews after the closest dates to interviews and at least one assessment before the closest dates to interviews for the stable future trajectory modeling. For our cross-sectional correlation analysis linking language-based assessments with PTSD, we selected PCL scores of WTC responders as their cross-sectional PTSD symptom severity at the time of the interview (Interview PCL), and it is controlled for future PTSD trajectories as a baseline.

*Trajectory Prediction*

For modelling the trajectory of PCL scores of each responder, we fit an ordinary least squares regression model with an intercept to the post-interview PCL scores as a function of time t:

$$PCL_{it} = \beta_{0i} + \beta_{1i}t + \varepsilon_{it}^{(t)} \qquad (1)$$

where PCL scores were measured at (t) years after the interviews then use the $\beta_{1i}$ coefficient as a future PCL score trajectory of a responder (i). Then, for the person-level prediction over $\beta_{1i}$ using the language based assessments controlling the age, gender, occupation, and years between the interview and 9/11 of the responder i as following:

$$\beta_{1i} = \alpha_0 + \alpha_1 x_{1i} + \alpha_2 x_{2i} + ... + \alpha_6 x_{6i} + \varepsilon_i^{(i)} \qquad (2)$$

where $x_1$: language based assessments, $x_2$: baseline PCL, $x_{3...6}$: age, gender, occupation, years after 9/11 (all variables standardized). Using equation (1) and (2), we use the following joint model:

$$PCL_{it} = \beta_{0i} + (\alpha_0 + \alpha_1 x_{1i} + \alpha_2 x_{2i} + ... + \alpha_5 x_{5i})t + \varepsilon_{it}^{(t)} \qquad (3)$$

and evaluate an effect size of each language-based assessment as its predictive power for future PCL trajectories of the responders.

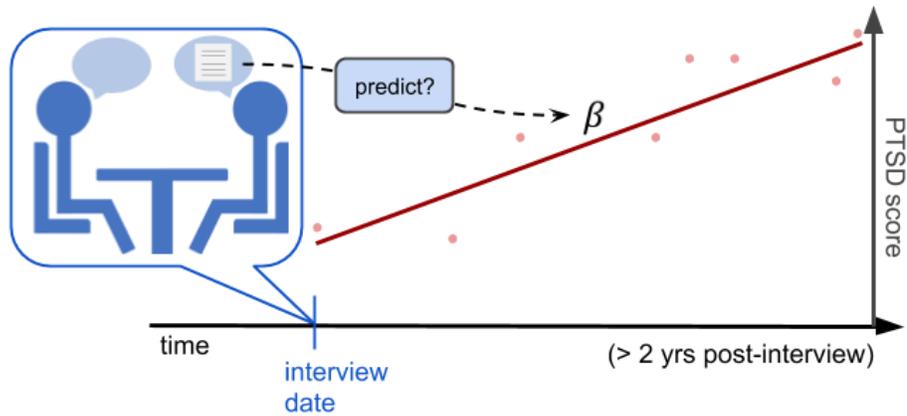

**Fig. 1**. Evaluation setup for trajectory prediction. According to equation 3, we can then model the control-adjusted trajectory per user as $B_{1\text{-cntrl}, i} = (\alpha_0 + \alpha_1 x_{1i} + \alpha_2 x_{2i} + ... + \alpha_5 x_{5i})$. Then, we used the slope of the fitted line as the PCL trajectory of the corresponding subject. Our main outcome was correlations between this trajectory slope and the subject's language patterns.

**Results**

Most responders were male (90%) and half (48%) were police (see Table 1 for sample characteristics). Their median age at their interviews was 55 (53 for the longitudinal cohort). The median number of words across the interviews was 10,254.

*Associations between language-based assessments and mental health variables*

Table 2 shows the linear regression analyses linking language-based assessments and PCL scores among responders around their interview dates. Higher PCL scores were significantly associated with language-based assessments consistent with anxious, depressive, and neuroticism. High scores were also associated with greater use of first-person singular and more total count of words in their interviews ($r >$ 0.22). Conversely, higher scores were also associated with less extraversion language patterns, first-person plurals, and articles. Results remained unchanged after adjusting for age, gender, occupation, and years after 9/11 despite some effects from covariates (<0.07).

**Table 2.** *Cross-sectional association between language-based assessments and PCL PTSD Score.*

| | Interview Language Features | PTSD Symptoms | |
|---|---|---|---|
| | | *r* (direct correlation with symptom slope) | *β* (adjusted for age, gender, occupation, yrs since 9-11) |
| **Psychological Traits** | Anxiety | 0.26 [0.03, 0.46] | 0.19 [-0.04, 0.40] |
| | Depression | 0.38* [0.16, 0.56] | 0.32* [0.09, 0.51] |
| | Neuroticism | 0.32* [0.10, 0.51] | 0.25 [0.02, 0.45] |
| | Extraversion | -0.10 [-0.32, 0.13] | -0.15 [-0.37, 0.08] |
| **Linguistic Style** | First-Person Singular | 0.31* [0.09, 0.50] | 0.31* [0.09, 0.51] |
| | First-Person Plural | -0.05 [-0.28, 0.18] | -0.09 [-0.31, 0.14] |
| | Articles | -0.09 [-0.31, 0.14] | -0.04 [-0.27, 0.19] |
| | AVG Word Length | 0.05 [-0.18, 0.27] | 0.03 [-0.20, 0.26] |
| | Word Count | 0.22 [-0.01, 0.43] | 0.16 [-0.07, 0.38] |

Associations are from ordinary least squares over standardized independent variable -- the language based assessment and the standardized dependent variable -- PTSD Checklist scores (PCL scores). Without controls is equivalent to Pearson Product-Moment Correlation (N=75). Square brackets indicate 95% confidence intervals. Controls included as covariates (right column) included age, gender, occupation, years between 9/11/01 and interview date. * indicates significant correlations (multi-test, Benjamini-Hochburg adjusted $p < 0.050$). Each row is color-coded separately, from red (negative correlations) to green (positive correlations); greyed values indicate non-significant.

*Trajectory analysis*

Table 3 shows that language-based assessments of the oral histories significantly predicted responders' PCL trajectories during the follow-up period. First, we calculated linear regression coefficient effect sizes when we modeled PCL score trajectories with language features only (first column of Table 3). Then we add the control variables into the model (the second column). Although the general directions of correlations were the same both with and without controls, suppression effects of control variables increased the effect sizes for anxiety and first-person plural usage.

**Table 3.** Predicting PCL trajectories of the responders using language-based assessments

|  | Interview Language Features | PTSD Symptoms Future Trajectories | |
|---|---|---|---|
|  |  | *r* (direct correlation with symptom slope) | *β* (adjusted for age, gender, occupation, yrs since 9-11, Interview PCL) |
| **Psychological Traits** | Anxiety | 0.16 [-0.07, 0.37] | 0.31* [0.09, 0.50] |
|  | Depression | -0.00 [-0.23, 0.22] | 0.17 [0.06, 0.38] |
|  | Neuroticism | 0.07 [0.29, -0.16] | 0.21 [0.42, -0.01] |
|  | Extraversion | 0.17 [-0.06, 0.38] | 0.18 [-0.05, 0.39] |
| **Linguistic Style** | First-Person Singular | 0.00 [-0.23, 0.23] | 0.10 [-0.13, 0.32] |
|  | First-Person Plural | -0.36* [-0.54, -0.14] | -0.37* [-0.55, -0.16] |
|  | Articles | -0.16 [-0.37, 0.07] | -0.25 [-0.45, -0.02] |
|  | AVG Word Length | -0.36* [-0.54, -0.14] | -0.36* [-0.54, -0.14] |
|  | Word Count | 0.06 [-0.17, 0.28] | 0.19 [-0.04, 0.40] |

Associations are from ordinary least squares over standardized independent variable -- the language based assessment and the standardized dependent variable -- PCL future trajectory. Without controls is equivalent to Pearson Product-Moment Correlation (N=75) with controls: age, gender, occupation, interview years since 9/11, and interview PCL score.

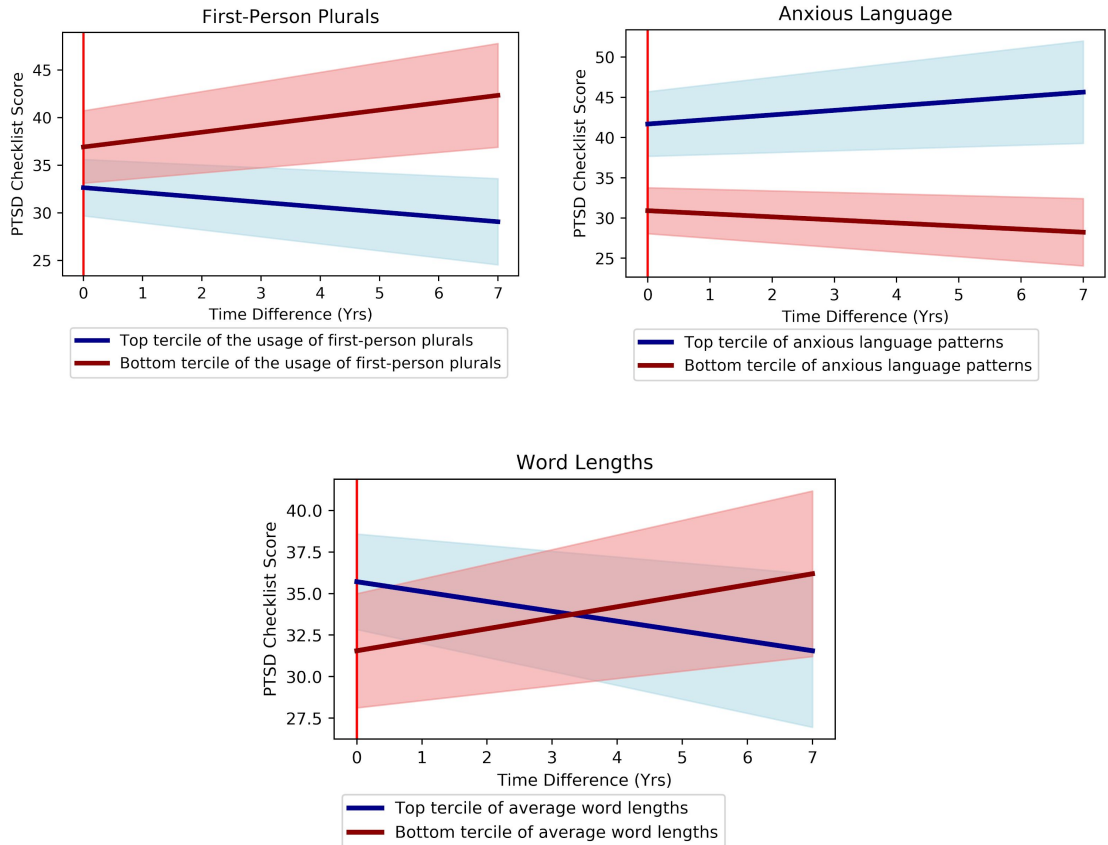

**Fig. 2**. Average future PCL score trajectories of top (red) and bottom (blue) terciles of responders based on language-based assessments: word usages of first-person plurals (left), anxious language patterns (right) and average word lengths (bottom). All trajectories have been adjusted for interview (baseline) PCL scores, representing the residual after accounting for the expected trajectory at baseline. All differences are significant at p<0.05 (see supplemental table S1 for further analysis).

**Discussion**

The goal of the present study was to examine whether AI-based language assessments developed in non-clinical contexts were reliably 1) associated with PTSD on self-reported questionnaires, and 2) able to predict the extent to which one's symptoms would get better or worse (trajectory) within a long-term clinical setting. The study found support for the view that language-based assessments could be reliably used in a clinical setting when processing naturalistic interviews: specifically, we found that language-based features were indicative of current functioning (supporting aim 1) and that language-based features could predict future PTSD symptom trajectories (supporting aim 2). This study, for the first time, suggested that AI assessments of interviews from a clinical sample not focused specifically on the topic of mental health could be used to identify features indicative of a person's current and future mental well-being.

*Implications*

There are three major implications from this work. First, AI-based assessments of interviews were associated with assessment of mental health scores concurrently, supporting the first aim. Specifically, depressed language was associated with greater PTSD symptom severity, as is self-focused language. This corroborates the clinical conceptualization of PTSD as involving self-focused rumination which maintains PTSD symptoms over time (Michael, Halligan, Clark, & Ehlers, 2007).

Second, use of more anxious language predicted increased PTSD symptoms in the future, even when adjusting for age, gender, occupation, and years since the 9/11 disaster. This suggests that while immediate PTSD severity is associated with low mood, a worsening of PTSD is determined by anxiety, rather than depression. These results may suggest that while immediate PTSD severity is reflected in affective experience, it may be the cognitive processes associated with anxiety (worry, rumination) that underlie future increases in PTSD symptoms. This dovetails with accounts of PTSD that understand it to be maintained through rumination and worry (Michael et al., 2007).

Third, use of more first-person plural pronouns ("we", "us", "our") predicted decreased PTSD symptoms in the future when adjusting for the confound variables. This supports research showing that social support is an important affordance that can buffer against and help alleviate the psychopathological load of a traumatic life event. Previous findings have suggested that processes associated with chronic sympathetic arousal (which include the chronic activation of the HPA-axis in states of hypervigilance) may be "buffered against" by social interactions with kin and close others (e.g., McGowan, 2002).

*Depressive Language and Current PTSD Severity*

Depressive language ($\beta$=0.32; p=0.044) and high usage of first-person singulars ($\beta$=0.31; p=0.044) were most highly correlated with high PCL scores even after accounting for age, gender, years since 9/11, and responder occupation. These associations were consistent with findings from prior studies showing an association of PTSD symptoms with increased risk of depression (Breslau, Davis, Peterson, & Schultz, 2000; Stander, Thomsen, & Highfill-McRoy, 2014). Similarly, high usage of first-person singular in messages are negatively correlated with life satisfaction (Schwartz et al., 2013). Anxious and neurotic language patterns had strong positive correlations with PCL scores, which align with a previous study that identified avoidance and hyperarousal symptoms as frequently reported symptoms (Bromet et al., 2016). For the associations between personality traits and PTSD symptoms, previous studies found that low extraversion and high neuroticism are associated with an increased risk of PTSD (Breslau, Davis, & Andreski, 1995; Fauerbach, Lawrence, Schmidt, Munster, & Costa, 2000), and we observed the same patterns of our language-based extraversion and neuroticism with PTSD severity.

*Predictors of PTSD Symptom Trajectories*

We examined language-based assessments as a predictor of responders' PCL trajectories after their interviews. Usage of first-person plurals and longer average word lengths were most highly correlated with improvement in PTSD in all cases, whether adjusting for baseline PCL-score and demographics or not. For other language-based assessments, coefficient effect sizes increased when we accounted for confounding due to the suppression effects mainly attributable to PCL scores, age at interview, and gender (see supplemental Table S1).

Furthermore, we analyzed potential mediation effects of marital status to address whether differences in the use of pronouns were merely reflecting marital status although previous work does not suggest such a relationship (Simmons, Gordon, & Chambless, 2015). Our results showed that these two types of language-based assessments predicted beyond marital status as their correlations remained statistically significant after adjusting for both controls and marital status (see supplemental Table S2 and Table S3). This demonstrates that these linguistic markers capture an orientation towards the self and others over and above marital status.

*Social Support*

In line with an extensive literature in psychology, we observed the use of "I" vs. "we" pronouns to mark classes of psychological processes that determined adjustment to and recovery from trauma. Previous work has related higher use of first-person singular pronouns ("I"-talk) self-focus (Carey et al., 2015); we found it correlated with high cross-sectional PTSD severity. On the other hand, we found high usage of first-person plural pronouns ("we") to be associated with a decrease of PTSD symptoms in the future. **Self-focused thinkin**g has been identified as a transdiagnostic factor of PTSD and depressive symptoms marking an often maladaptive preoccupation with the self and negative experience (Martin, 1985; Ingram, 1990; Birrer & Michael, 2011). The use of "I" pronouns, in turn, has previously been found to be a dependable marker of self-focus in natural language (Wegner & Giuliano 1980; Watkins and Teasdale, 2001; Carey et al., 2015). Beyond mere self-focus, depression and negative affectivity have also been robustly associated with higher use first person singular pronouns (Rude et al., 2004; Holtzman, 2017) and PTSD (Miragoli, Camisasca, & Di Blasio, 2019); PTSD also with few "we" pronouns (Papini, Yoon, Rubin, Lopez-Castro, & Hien, 2015). Our study showed further evidence for these patterns: greater use of "I" pronouns positively correlated with severe cross-sectional PTSD symptoms , and high usage of "we" pronouns predicted decreasing PTSD symptoms in the future.

*Limitations*

This was the first study to evaluate the relationship between automatic language-based assessments from interviews and PTSD symptoms of a trauma population, and there were several limitations. First, our sample covered a particular cohort of trauma survivors, those responding to the WTC disaster, so generalizability of present findings to other occupations and demographic groups is unknown. Future research would need to investigate whether the results replicate to additional populations. Second, language-based assessment predicted future change in PTSD and suggested that cognitive and social risk processes may be involved, but mechanisms underpinning these predictive effects were not tested directly. Third, while our feature-based identification process was completed in a large database with ample capacity to train robust AI models, the present analysis had a relatively small sample size that could only be reliably used for application and was too small to retrain models for the current population. Future work in larger samples will be able to tailor AI-based assessments to specific populations and clinical questions substantially enhancing their predictive power.

*Potential use in Clinical Care*

Clinical evaluation of PTSD symptoms in trauma-exposed patients is time-consuming and burdensome. Moreover, primary care providers often lack expertise to complete this assessment. Our results show that natural language can provide clinically useful information both for detection of PTSD and prediction of future symptom escalation. These methods can be applied to routine clinical interviews completed by staff without mental health expertise. Although oral history interviews used in this project were lengthy, previous research has shown that interactions as brief as 5 minutes (e.g. 200 words) can be sufficient to obtain reliable AI-based assessments (Kern et al., 2016). These assessments would not replace a psychiatric evaluation, but can be useful for screening in primary care and as an aid to psychiatrists, picking-up on diagnostic and prognostic features in language that may be missed clinically. Specific language-based risk factors could inform treatment selection, such as low social support, and may suggest group therapy or peer support interventions, whereas maladaptive cognitive styles suggest cognitive behavioral therapy.

*Conclusion*

We found automated AI-based assessments utilizing the language of WTC responders in their oral history interviews predicted their PTSD symptoms in both cross-sectional and longitudinal trajectory analyses. The patterns and the correlations from these studies should be examined cautiously, and may require independent confirmations from other WTC cohorts and across different types of exposures before general applications for PTSD treatments. Still, the patterns of language-based assessments consistent with previous findings in other settings and their strong statistical correlations provided unique insights and explanations beyond commonly known confounds or risk factors such as age, gender, occupation, marital status or even questionnaire-based depression measures, suggesting support for clinicians towards more precise decisions. More generally, language-based assessments which capture individual digital phenotypes and distinctive linguistic markers from transcripts of interviews are very useful for investigating underlying causes of PTSD and may play a critical role as a supplement for enhancing personalized preventive care (Hamburg & Collins, 2010) and more effective treatments for PTSD; they may even enable real-time screening or preventive measures with reduced costs and less therapist time for helping a large number of people exposed to large-scale traumatic events (e.g., natural disasters, WTC PTSD) similar to a previous online PTSD treatment (Lewis et al., 2017). Nevertheless, future studies with applying language-based assessment on larger samples will be required in order to more precisely validate their statistical significance and correlations, and even further studies into subphenotypes and more detailed categorizations of language-based assessments will lead to more diverse analysis with rich high-dimensional digital phenotypes.

## Acknowledgements

The authors are extremely grateful to the WTC rescue and recovery workers, who gave of themselves so readily in response to the WTC attacks and agreed to participate in this ongoing research effort. We also thank the clinical staff of the World Trade Center Medical Monitoring and Treatment Programs for their dedication and the labor and community organizations for their continued support. Son and Schwartz were supported, in part, by NIH R01 AA028032-01.

## Declaration of Interest

None

# Supplementary Materials

**Table S1**. Analysis on suppression effects of confound variables for PCL trajectories

| | | **PTSD Symptoms Trajectories** (Future Change) | | | | | | |
|---|---|---|---|---|---|---|---|---|
| | | r [95% CI] | β (adjusted for each confound variable) [95% CI] | | | | | |
| | **Interview Language Features** | Unadjusted | Occupation | Gender | Years since 9/11 | Interview PCL | Age | Adjusted[1] |
| **Psychological Traits** | Anxiety | 0.16 [-0.07, 0.37] | 0.17 [-0.06, 0.38] | 0.21 [-0.02, 0.41] | 0.16 [-0.07, 0.37] | 0.25 [0.03, 0.45] | 0.21 [-0.02, 0.41] | 0.31* [0.09, 0.50] |
| | Depression | 0.00 [-0.23, 0.22] | 0.00 [-0.22, 0.23] | 0.05 [-0.18, 0.27] | 0.00 [-0.23, 0.23] | 0.13 [-0.10, 0.35] | 0.03 [-0.20, 0.25] | 0.17 [0.06, 0.38] |
| | Neuroticism | 0.07 [0.29, -0.16] | 0.07 [-0.16, 0.29] | 0.12 [-0.11, 0.34] | 0.07 [-0.16, 0.29] | 0.18 [-0.05, 0.39] | 0.09 [-0.14, 0.31] | 0.21 [0.42, -0.01] |
| | Extraversion | 0.17 [-0.06, 0.38] | 0.17 [-0.06, 0.38] | 0.21 [-0.01, 0.42] | 0.17 [-0.06, -0.38] | 0.15 [-0.09, 0.36] | 0.18 [-0.05, 0.39] | 0.18 [-0.05, 0.39] |
| **Linguistic Style** | First-Person Singular | 0.00 [-0.23, 0.23] | -0.01 [-0.24, 0.22] | 0.01 [-0.22, 0.24] | 0.00 [-0.23, 0.23] | 0.10 [-0.13, 0.32] | 0.02 [-0.21, 0.25] | 0.10 [-0.13, 0.32] |
| | First-Person Plural | -0.36* [-0.54, -0.14] | -0.36* [-0.54, -0.14] | -0.33* [-0.52, -0.11] | -0.36* [-0.54, -0.14] | -0.37* [-0.55, -0.16] | -0.37* [-0.55, -0.16] | -0.37* [-0.55, -0.16] |
| | Articles | -0.16 [-0.37, 0.07] | -0.19 [-0.40, 0.04] | -0.19 [-0.40, 0.04] | -0.16 [-0.38, 0.07] | -0.19 [-0.40, 0.04] | -0.19 [-0.40, 0.04] | -0.25 [-0.45, -0.02] |
| | AVG Word Length | -0.36* [-0.54, -0.14] | -0.37* [0.55, -0.15] | -0.35* [-0.53, -0.13] | -0.37* [-0.55, -0.15] | -0.34* [-0.53, -0.13] | -0.36* [-0.54, -0.14] | -0.36* [-0.54, -0.14] |
| | Word Count | 0.06 [-0.17, 0.28] | 0.07 [-0.16, 0.29] | 0.10 [-0.13, 0.32] | 0.06 [-0.17, 0.29] | 0.13 [-0.10, 0.35] | 0.09 [-0.14, 0.31] | 0.19 [-0.04, 0.40] |

*p<0.05, Benjamini-Hochberg (False-Discovery Rate) corrected.
[1]Adjusted for occupation, gender, years since 9/11, Interview PCL, and age

**Table S2**. Effects of marital status as a confound variable for cross-sectional interview PCL scores

| | | PTSD Symptoms (concurrent, cross-sectional) | | | |
|---|---|---|---|---|---|
| | | *r* [95% CI] | *β* (adjusted for each confound variable) [95% CI] | | |
| | **Interview Language Features** | Unadjusted | Marital Status | Adjusted[1] | Adjusted[2] |
| **Psychological Traits** | Anxiety | 0.26 [0.03, 0.46] | 0.25 [0.02, 0.45] | 0.19 [-0.04, 0.40] | 0.19 [-0.04, 0.39] |
| | Depression | 0.38* [0.16, 0.56] | 0.37* [0.16, 0.55] | 0.32* [0.09, 0.51] | 0.31* [0.09, 0.51] |
| | Neuroticism | 0.32* [0.10, 0.51] | 0.31* [0.09, 0.50] | 0.25 [0.02, 0.45] | 0.25 [0.03, 0.45] |
| | Extraversion | -0.10 [-0.32, 0.13] | -0.09 [-0.31, 0.14] | -0.15 [-0.37, 0.08] | -0.15 [-0.36, 0.08] |
| **Linguistic Style** | First-Person Singular | 0.31* [0.09, 0.50] | 0.32* [0.10, 0.51] | 0.31* [0.09, 0.51] | 0.32* [0.10, 0.51] |
| | First-Person Plural | -0.05 [-0.28, 0.18] | -0.06 [-0.28, 0.17] | -0.09 [-0.31, 0.14] | -0.09 [-0.31, 0.14] |
| | Articles | -0.09 [-0.31, 0.14] | -0.08 [-0.30, 0.15] | -0.04 [-0.27, 0.19] | -0.03 [-0.26, 0.19] |
| | AVG Word Length | 0.05 [-0.18, 0.27] | 0.06 [-0.17, 0.28] | 0.03 [-0.20, 0.26] | 0.03 [-0.20, 0.26] |
| | Word Count | 0.22 [-0.01, 0.43] | 0.20 [-0.03, 0.41] | 0.16 [-0.07, 0.38] | 0.15 [-0.08, 0.37] |

[1] Adjusted for age, gender, occupation, and years since 9/11.
[2] Adjusted for age, gender, occupation, years since 9/11, and marital status.
*p<0.05, Benjamini-Hochberg (False-Discovery Rate) corrected.

**Table S3**. Slight mediation effects of marital status for PCL score future trajectories

| | | PTSD Symptoms Trajectories (Future Change) | | | |
|---|---|---|---|---|---|
| | | *r* [95% CI] | β (adjusted for confound variables) [95% CI] | | |
| | **Interview Language Features** | Unadjusted | Adjusted[1] | Adjusted[2] | Adjusted[3] |
| **Psychological Traits** | Anxiety | 0.16 [-0.07, 0.37] | 0.25 [0.02, 0.45] | 0.31* [0.09, 0.50] | 0.31* [0.09, 0.50] |
| | Depression | -0.00 [-0.23, 0.22] | 0.13 [-0.10, 0.35] | 0.17 [0.06, 0.38] | 0.17 [-0.06, 0.38] |
| | Neuroticism | 0.07 [0.29, -0.16] | 0.18 [-0.05, 0.39] | 0.21 [0.42, -0.01] | 0.22 [-0.01, 0.43] |
| | Extraversion | 0.17 [-0.06, 0.38] | 0.15 [-0.08, 0.36] | 0.18 [-0.05, 0.39] | 0.19 [-0.04, 0.40] |
| **Linguistic Style** | First-Person Singular | 0.00 [-0.23, 0.23] | 0.11 [-0.12, 0.33] | 0.10 [-0.13, 0.32] | 0.11 [-0.12, 0.33] |
| | First-Person Plural | -0.36* [-0.54, -0.14] | -0.38* [-0.56, -0.16] | -0.37* [-0.55, -0.16] | -0.37* [-0.55, -0.16] |
| | Articles | -0.16 [-0.37, 0.07] | -0.19 [-0.40, 0.04] | -0.25 [-0.45, -0.02] | -0.24 [-0.44, -0.02] |
| | AVG Word Length | -0.36* [-0.54, -0.14] | -0.34* [-0.53, -0.12] | -0.36* [-0.54, -0.14] | -0.35* [-0.54, -0.14] |
| | Word Count | 0.06 [-0.17, 0.28] | 0.13 [-0.10, 0.34] | 0.19 [-0.04, 0.40] | 0.18 [-0.05, 0.39] |

[1] Adjusted for Marital Status and Interview PCL
[2] Adjusted for age, gender, occupation, years since 9/11, and Interview PCL score.
[3] Adjusted for age, gender, occupation, years since 9/11, interview PCL score, and marital status.
*p<0.05, Benjamini-Hochberg (False-Discovery Rate) corrected.